\documentclass[conference]{IEEEtran}
\usepackage{amsmath}
\usepackage{graphicx}
\usepackage{times}
\usepackage{booktabs}
\usepackage{graphicx} % for \includegraphics
\usepackage{subcaption} % for subfigures
\usepackage{amsfonts}
\usepackage{caption} % for better captions
\usepackage{lipsum}
\usepackage[numbers]{natbib}
\usepackage{multicol}
\usepackage[bookmarks=true, hidelinks]{hyperref}
\usepackage[T1]{fontenc}
\usepackage[utf8]{inputenc}
\usepackage{pmboxdraw}
\usepackage{verbatim}

\IEEEoverridecommandlockouts

\begin{document}

\title{LunarLoc: Segment-Based Global Localization \\ on the Moon}

% You will get a Paper-ID when submitting a pdf file to the conference system
% \author{Author Names Omitted for Anonymous Review}
% \author{Annika Thomas, Robaire Galliath, Aleksander Garbuz, Luke Anger, Cormac O'Neill, \\ Trevor Johst, Dami Thomas, George Lordos, Jonathan P. How}

\author{
\IEEEauthorblockN{Annika Thomas, Robaire Galliath, Aleksander Garbuz, Luke Anger, Cormac O'Neill, \\ Trevor Johst, Dami Thomas, George Lordos, Jonathan P. How}

\thanks{Authors are affiliated with Massachusetts Institute of Technology, Cambridge, MA 02139, USA. \texttt{annikat@mit.edu}.}
\thanks{The simulation used for this research was developed as part of the Lunar Autonomy Challenge. The Lunar Autonomy Challenge is a collaboration between NASA, The Johns Hopkins University (JHU) Applied Physics Laboratory (APL), Caterpillar Inc., and Embodied AI. APL is managing the challenge for NASA.}
}

%\author{\authorblockN{Michael Shell}
%\authorblockA{School of Electrical and\\Computer Engineering\\
%Georgia Institute of Technology\\
%Atlanta, Georgia 30332--0250\\
%Email: mshell@ece.gatech.edu}
%\and
%\authorblockA{Twentieth Century Fox\\
%\and
%\authorblockN{James Kirk\\ and Montgomery Scott}
%\authorblockA{Starfleet Academy\\
%San Francisco, California 96678-2391\\

% avoiding spaces at the end of the author lines is not a problem with
% conference papers because we don't use \thanks or \IEEEmembership

%Homer Simpson\authorrefmark{2},
%Montgomery Scott\authorrefmark{3} and
%Eldon Tyrell\authorrefmark{4}}
%Telephone: (800) 555--1212, Fax: (888) 555--1212}
%\authorblockA{\authorrefmark{4}Tyrell Inc., 123 Replicant Street, Los Angeles, California 90210--4321}}

\maketitle

\begin{abstract}
Global localization is necessary for autonomous operations on the lunar surface where traditional Earth-based navigation infrastructure, such as GPS, is unavailable. As NASA advances toward sustained lunar presence under the Artemis program, autonomous operations will be an essential component of tasks such as robotic exploration and infrastructure deployment. Tasks such as excavation and transport of regolith require precise pose estimation, but proposed approaches such as visual-inertial odometry (VIO) accumulate odometry drift over long traverses. Precise pose estimation is particularly important for upcoming missions such as the ISRU Pilot Excavator (IPEx) that rely on autonomous agents to operate over extended timescales and varied terrain. To help overcome odometry drift over long traverses, we propose LunarLoc, an approach to global localization that leverages instance segmentation for zero-shot extraction of boulder landmarks from onboard stereo imagery. Segment detections are used to construct a graph-based representation of the terrain, which is then aligned with a reference map of the environment captured during a previous session using graph-theoretic data association. This method enables accurate and drift-free global localization in visually ambiguous settings. LunarLoc achieves sub-cm level accuracy in multi-session global localization experiments, significantly outperforming the state of the art in lunar global localization. To encourage the development of further methods for global localization on the Moon, we release our datasets publicly with a playback module: \url{https://github.com/mit-acl/lunarloc-data}.
\end{abstract}

\IEEEpeerreviewmaketitle

\section{Introduction}
Autonomous systems will play a vital role in preparing humanity for a sustained presence on the Moon. Precise and robust global localization will be a key capability for enabling autonomous surface operations to operate drift-free during precision-based tasks. The ability for robotic systems to accurately determine their position and orientation within a shared lunar reference frame is critical for navigation, exploration, infrastructure deployment, and scientific discovery. Unlike Earth-based systems that benefit from Global Navigation Satellite Systems (GNSS), the Moon lacks robust GNSS support, requiring global localization techniques that integrate on-board perception and environmental understanding for accurate mapping.

\begin{figure}[t]
    \centering
    \includegraphics[width=\linewidth, trim={4.5cm, 4cm, 6cm, 0}, clip]{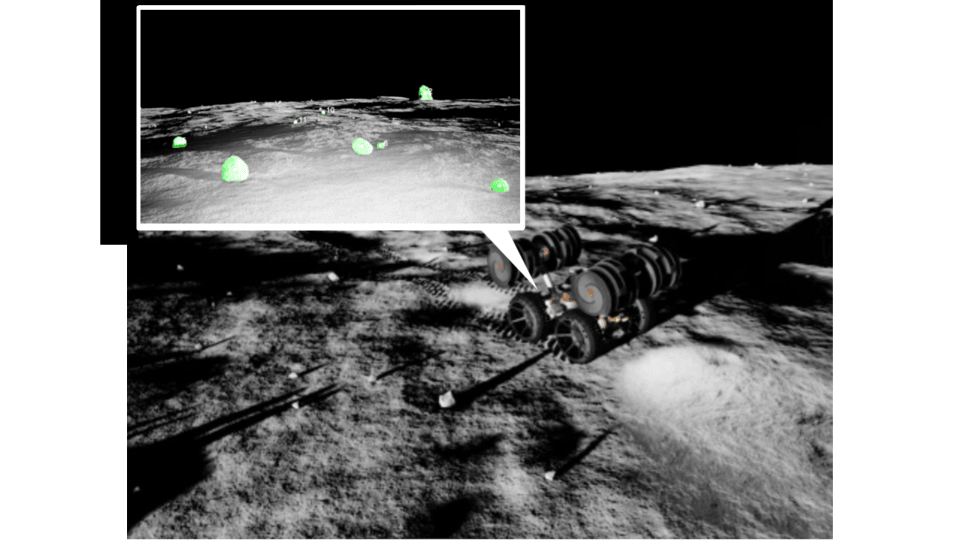}
    \caption{LunarLoc uses open-set segmentation to generate maps through detection of boulders (shown in green), then performs object-based global localization using multi-session maps.}
    \label{fig:enter-label}
\end{figure}

This ability for robots to localize without dependence on GNSS is particularly challenging in the context of lunar operations where harsh lighting conditions drastically alter the appearance of the environment. Classical approaches for localization involve visual place recognition \cite{arandjelovic2016netvlad, oquab2023dinov2} followed by visual feature extraction \cite{sift2004, orb2011} and matching \cite{fischler1981random}. Place recognition methods struggle with viewpoint variation and may fail to provide descriptors that are different enough for meaningful association across places \cite{10801471}. While feature-based approaches are effective for applications such as frame-to-frame tracking, they may fail to provide meaningful visual cues for long-term localization. These limitations of current global localization methods motivate the need for a separate approach to global localization on the Moon.

LunarLoc uses the underlying structure of the environment as a cue for global localization. The lunar environment has an abundance of rocks, the positions of which do not change under varying lighting conditions. Zero-shot extraction of segments within the scene provides consistent detections of rocks regardless of lighting conditions. LunarLoc provides a robust paradigm for global localization on the lunar surface, as demonstrated through experiments with a digital twin of the IPEx rover operating in a simulated lunar environment.

Our contributions include:
\begin{itemize}
    \item Real-time object-based mapping using open-set segmentation for multi-session graph-theoretic global localization on the lunar surface
    \item Experimental validation of our approach in a simulated lunar environment, achieving cm-level accuracy which significantly outperforms the state of the art for lunar global localization
    \item Release of our datasets as well as a simulation playback environment to encourage further development of lunar global localization
\end{itemize}

\section{Related Works}

\begin{figure*}[h!]
    \centering
    % First subfigure
    \begin{subfigure}{0.49\textwidth}
        \includegraphics[width=\linewidth, trim={0, 1.4cm, 0.2cm, 0.75cm}, clip]{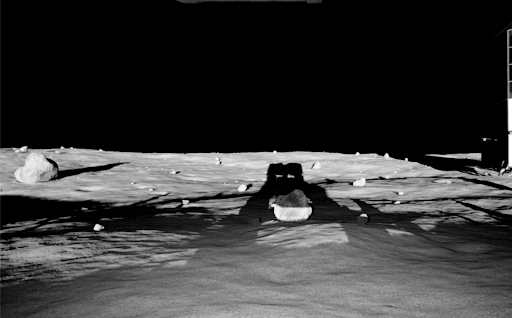}
        % \caption{First}
        \label{fig:sub1segment}
    \end{subfigure}
    \hfill
    % Second subfigure
    \begin{subfigure}{0.49\textwidth}
        \includegraphics[width=\linewidth, trim={0, 0, 0, 0.6cm}, clip]{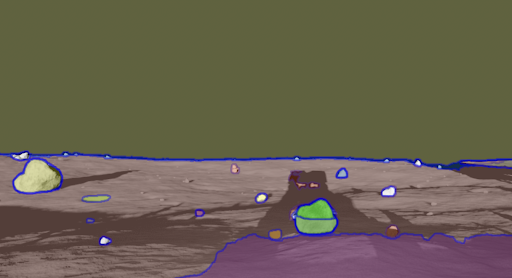}
        % \caption{Second}
        \label{fig:sub2}
    \end{subfigure}

    \caption{Extraction of segment locations via open-set segmentation with FastSAM \cite{zhao2023fast}. Images include a front-view image taken from the IPEx digital twin in the Lunar Simulator (left) and the zero-shot segmented image (right).}
    % \caption{We perform open-set segmentation using FastSAM to extract the segment locations from images. Images include a front-view image taken from the Lunar Simulator front camera of the IPeX (left) and the zero-shot segmented image (right). }
    \label{fig:segment}
\end{figure*}

% \begin{table}[h!]
% \centering
% \caption{Planetary Localization Methods}
% \begin{tabular}{@{}ccccc@{}}
% \toprule
% \textbf{Modality} & \textbf{Method} & \textbf{Locale} & \textbf{Accuracy [m]} & \textbf{Demonstration} \\ \midrule

% Local Imagery & LunarLoc & Moon & 0.02 & LunarLoc \\
% Satellite Time Code & GPS & Earth & 4.9 & Various \cite{vanDiggelen2015gpsmooc} \\
% Satellite Imagery & Geo-Referencing & Mars & 0.5 & Spirit / Opportunity \cite{parker2010geomorphic} \\
% Local Imagery & TRN \cite{johnson2007general} & Mars & 10 & Perseverance \cite{hook2022topographical} \\
% Aerial Imagery & MAVeN \cite{martin2017maven} & Mars & 3 & Ingenuity \cite{grip2022ingenuity} \\
% Local Imagery & LunarNav \cite{daftry2023lunarnav} & Moon & 5 & - \\
% Local Imagery & ShadowNav \cite{atha2024shadownav} & Moon & 1.5 & - \\

% \bottomrule
% \label{tab:planetary-localization}
% \end{tabular}
% \end{table}

Several approaches have been proposed for global localization in environments without GNSS support and unstructured environments. We review these approaches as well as landmark-based localization and planetary localization.

\subsection{Global Localization in Environments without GNSS}

Image-based localization techniques are commonly used in environments with GNSS. Place recognition approaches such as bag of words \cite{cummins2008fab} and NetVLAD \cite{arandjelovic2016netvlad} store image representations as embeddings in a database which are used to associate new images with previous observations. These approaches are reasonably robust to changing environments from the same viewpoint.
Sarlin et. al \cite{sarlin2019coarse} built upon this concept to allow for online pose estimation using image-based localization. 
New images are compared to an existing library for an initial coarse estimation, before keypoint matching is applied for finer positioning. 
While these approaches demonstrate high recall in visually distinctive environments, their performance degrades when comparing different places that include similar features \cite{10801471}. Learning-based methods such as \cite{keetha2023anyloc} and \cite{detone2018superpoint} are also sensitive to perceptual aliasing in visually ambiguous settings. 
% Focus: Methods for absolute pose estimation without external positioning systems.

% Content: Techniques like image retrieval + pose regression, visual place recognition, topological vs metric localization, and relocalization in SLAM systems.

% Relevant work: FAB-MAP, NetVLAD, HF-Net, and hybrid systems combining global descriptors with local feature matching.

\subsection{Localization in Unstructured and Natural Environments}

Unstructured and natural environments present a challenging setting for many visual SLAM systems which may assume the presence of urban-centric fixtures. To address the lack of semantic information such as lane markings and buildings, \cite{grimes2009efficient} integrates wheel odometry with visual tracking and \cite{ort2018autonomous} includes topological maps in the localization process. Approaches such as \cite{ankenbauer2023global} and \cite{10801471} specifically use the underlying structure of the environment as a cue for global localization. 

% Focus: Challenges of perception and localization in environments lacking human-made structures or semantic cues.

% Content: Terrain-based navigation, object-based SLAM with rocks, trees, or other natural features; use of elevation maps or DEMs.

% Relevant work: Techniques from agricultural robotics, off-road navigation, planetary analog terrain tests, and natural scene understanding using segmentation or edge features.

\subsection{Landmark-Based and Object-Centric Localization}

To overcome issues arising from perceptual aliasing, recent works have proposed landmark-based and object-centric localization using graph theoretic alignment between objects within maps. These approaches are lightweight, lending themselves well to multi-agent scenarios as well. Ankenbauer et. al \cite{ankenbauer2023global} proposed object-centric global localization in urban and unstructured scenarios, relying on YOLO \cite{redmon2018yolov3} to extract objects from the environment which limits extension to settings without pre-determined object classes. To enable open-set operation, SOS-Match \cite{10801471} employs zero-shot segmentation from SAM to extract segments reflective of objects in the scene. ROMAN \cite{peterson2024roman} and \cite{singh2024open} incorporate semantics into the object-based formulation to further disambiguate between geometrically similar scenes.

\subsection{Planetary and Lunar Localization}
%Focus: Localization strategies specifically designed for planetary exploration missions, particularly Mars and the Moon.
%Content: Orbital-to-surface localization, use of satellite imagery, visual odometry, terrain-relative navigation (TRN), and feature matching across scales.
%Relevant work: NASA Mars rovers’ localization systems (e.g., visual odometry, sun tracking), TRN used in Mars 2020 / Perseverance, methods from iGAPS and other lunar mapping systems.

Due to the lack of GNSS coverage on both the Moon and Mars, previous missions have adopted several different strategies to perform localization. While most rovers utilize wheel odometry and inertial measurement units for relative localization, these methods are prone to odometry drift over longer traverses. The Spirit and Opportunity rovers overcame this by correlating local panoramas to satellite images at regular intervals \cite{parker2010geomorphic}. This method relies on high resolution images of the rover's operational site, and is highly prone to changes in lighting. Both additionally incorporated visual-odometry and celestial sensing, but the former is still prone to drift and the latter only provides accurate estimates for heading \cite{maimone2006surface}. For the task of landing, orbital images are again used as a reference against sensor data in Terrain Relative Navigation (TRN) \cite{johnson2007general}. This technique was later shown to provide 10 meter position accuracy on the Perseverance rover \cite{hook2022topographical}. A comparison of existing and proposed planetary localization methods is provided in Table \ref{tab:planetary-localization}. Of these methods, LunarLoc is the only approach that provides centimeter-level accuracy.

\begin{table}[t!]
\centering
\caption{Planetary Localization Methods}
\begin{tabular}{@{}lllc@{}}
\toprule
\textbf{Method} & \textbf{Modality} & \textbf{Locale} & \textbf{Accuracy [m]} \\ \midrule
GPS, \cite{vanDiggelen2015gpsmooc} & Satellite Time Code & Earth & 4.9 \\
Geo-Referencing \cite{parker2010geomorphic} & Satellite Imagery & Mars & 0.5\\
TRN \cite{johnson2007general, hook2022topographical} & Local Imagery & Mars & 10.0 \\
MAVeN \cite{martin2017maven, grip2022ingenuity} & Aerial Imagery & Mars & 3.0 \\
LunarNav \cite{daftry2023lunarnav} & Local Imagery & Moon & 5.0 \\
ShadowNav \cite{atha2024shadownav} & Local Imagery & Moon & 1.5 \\ 
\textbf{LunarLoc} (Proposed) & Local Imagery & Moon & 0.02 \\

\bottomrule
\label{tab:planetary-localization}
\end{tabular}
\end{table}

% Focus: Using semantic or geometric landmarks for global localization.

% Content: Object-based SLAM, semantic segmentation for scene understanding, boulder detection and tracking, zero-shot or few-shot generalization of object detectors.

% Relevant work: Scene graphs, graph matching algorithms, semantic map fusion, and works leveraging sparse but salient natural features.

% \subsection{Graph-Theoretic Approaches in SLAM and Localization}
% Focus: Representing environments and localization problems as graphs.

% Content: Graph matching, pose graph optimization, data association using graph similarity metrics, topological SLAM.

% Relevant work: DBoW2, OpenGV, and applications of spectral graph theory or GNNs for spatial reasoning.

\section{Methodology}

\subsection{Open-Set Object-Based Mapping}

Our mapping approach consists of performing zero-shot instance segmentation on input images using a pre-trained foundation model \cite{zhao2023fast}. The segments are primarily extracted as boulders in the scene, reflecting the underlying geometric structure of the environment as shown in Fig. \ref{fig:segment}. 

\begin{figure}[t]
    \centering
    \includegraphics[width=0.95\linewidth, trim={70, 15, 70, 15}, clip]{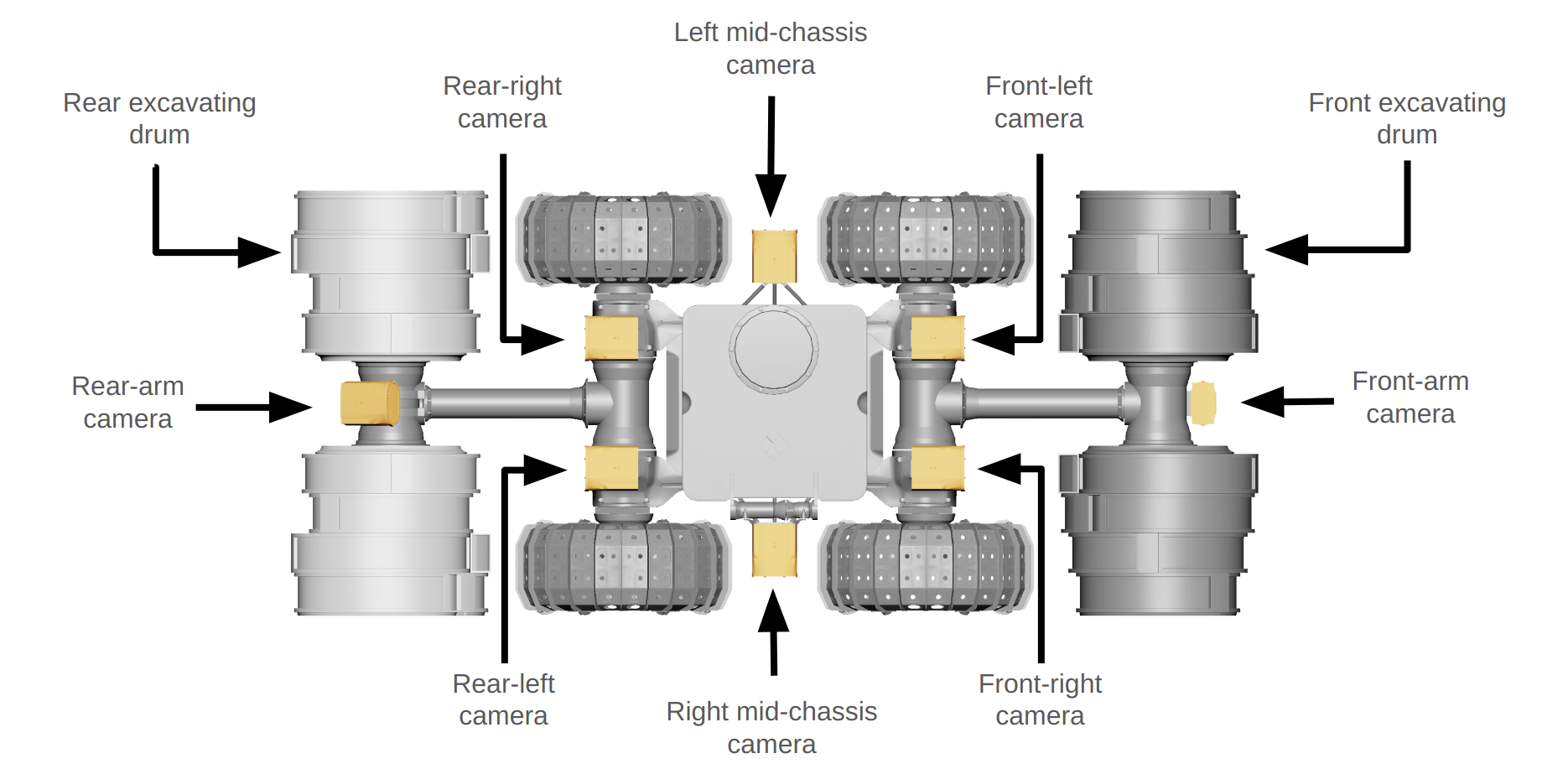}
    \caption{IPEx rover diagram where the front-left camera and rear-left camera are used as inputs to the segmentation model and front-right and rear-right cameras provide imagery for stereo depth estimation of mapped objects, from \cite{lunar_autonomy_challenge} }
    \label{fig:ipex-diagram}
\end{figure}

We perform mapping simultaneously from the front and back of the rover. Images captured from the front-left and rear-left IPeX rover cameras, as shown in Fig.~\ref{fig:ipex-diagram}, are used as inputs to the segmentation model.

We apply point prompting over a uniform $32 \times 32$ grid, producing a set of instance masks:
\begin{equation}
\mathcal{M} = \{ m_i \}_{i=1}^N,
\end{equation}
where each mask $m_i$ is defined by its image-frame centroid $\mathbf{c}_i \in \mathbb{R}^2$ and size $s_i \in \mathbb{R}^{+}$:
\begin{equation}
m_i = (\mathbf{c}_i, s_i).
\end{equation}

We find mask size by projecting each centroid $\mathbf{c}_i$ from the 2D image frame into 3D coordinates in the rover frame using stereo depth imagery. Assuming approximately spherical objects, we use the standard deviation $\sigma_i$ of the mask and the depth $D_i$ at $\mathbf{c}_i$ to estimate the physical size $s_i$ of the object:
\begin{equation}
s_i = \sigma_i \cdot D_i \cdot \alpha,
\end{equation}
where $\alpha$ is the angular resolution (e.g., radians per pixel) of the camera. This converts 2D pixel extent into a real-world spatial scale. 

We filter masks to remove those that (1) intersect the image boundary or (2) fall outside pre-defined size thresholds $s_{\text{min}} \leq s_i \leq s_{\text{max}}$. To further refine the set, we compute the 2D spatial covariance matrix of each mask and reject elongated shapes by enforcing an eigenvalue ratio constraint:
\begin{equation}
\frac{\lambda_{\text{max}}}{\lambda_{\text{min}}} \leq \tau_{\text{elong}},
\end{equation}
where $\lambda_{\text{max}}$ and $\lambda_{\text{min}}$ are the principal eigenvalues of the covariance matrix of the mask’s pixel distribution. Masks with high anisotropy are typically shadows or terrain edges and are excluded. The filtered set is denoted:
\begin{equation}
\widetilde{\mathcal{M}} = \left\{ m_i \in \mathcal{M} \ \middle|\ 
\begin{array}{l}
\text{$m_i$ is fully visible,} \\
s_{\text{min}} \leq s_i \leq s_{\text{max}}, \\
\lambda_{\text{max}} / \lambda_{\text{min}} \leq \tau_{\text{elong}}
\end{array}
\right\}.
\end{equation}

The final set of projected 3D centroids stored for each frame is:
\begin{equation}
M_{\text{veh}} = \{ \mathbf{C}_i \}_{i=1}^{\widetilde{N}}, \quad \mathbf{C}_i \in \mathbb{R}^3,
\end{equation}
where $\widetilde{N} = |\widetilde{\mathcal{M}}|$.

% We perform mapping simultaneously from the front and back of the rover. Images captured from the front-left and rear-left IPeX rover cameras as shown in Fig. \ref{fig:ipex-diagram} are used as inputs to the model.
% We use point prompting across a $32 \times 32$ grid of points to return a set of instance masks $\{m_{0}, m_{1}, ... m_{N}\}$. We then filter the masks to only include segments in full view of the camera, discarding those that intersect the border of the image or fall outside of minimum or maximum size thresholds. This results in a set of masks defined by their centroids in the image frame. Next, the centroid of each remaining mask is projected from the image frame to the rover frame using stereo imagery for depth. Assuming spherical objects, the standard deviation of each mask is used to estimate the true centroid. The resultant list of 3D object centroids, $M_{veh}$, is stored for each frame in the traverse. 

%If a given world-frame point has greater than $n$ detections across all centroids $\{S_0, ..., S_t\}$ after global registration, the cell is marked as occupied and included in the graph-theoretic association. 

The use of FastSAM \cite{zhao2023fast} allows this procedure to occur in real-time on a CPU, which is crucial for implementation of SLAM systems on the lunar surface. As the segmentation time is proportional to the prompt area, further speed-ups can additionally be achieved by making assumptions on the horizon line placement and excluding portions of the image.

\begin{figure}[b]
    \centering
    \begin{subfigure}{0.23\textwidth}
        \includegraphics[width=\linewidth, trim={3cm, 0, 5cm, 0}, clip]{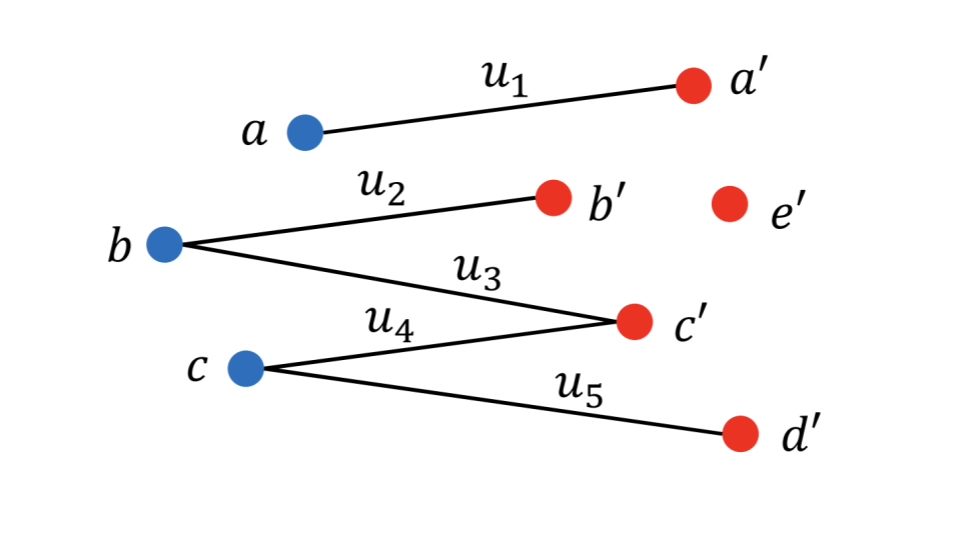}
        \caption{}
        \label{fig:sub1}
    \end{subfigure}
    \hfill
    \begin{subfigure}{0.23\textwidth}
        \includegraphics[width=\linewidth, trim={4cm, 0, 4cm, 0}, clip]{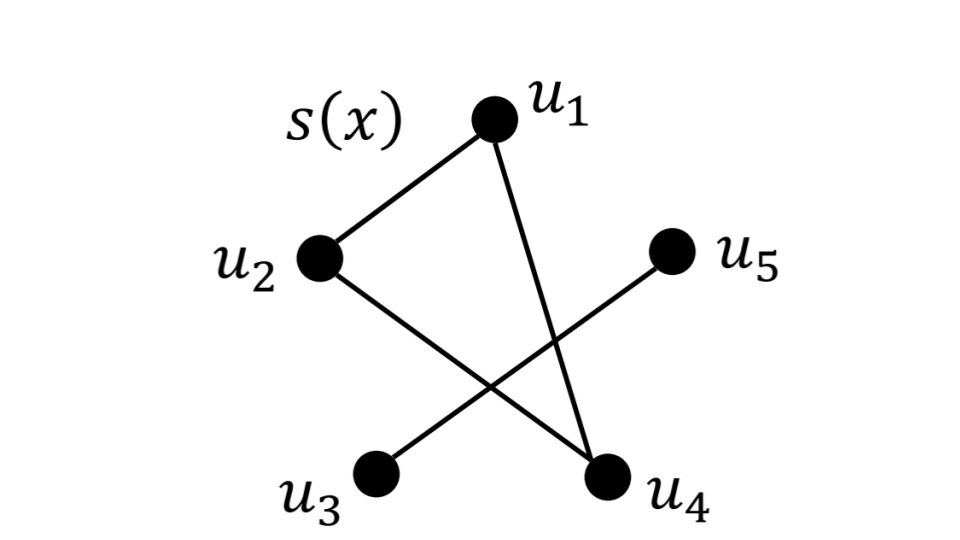}        
        \caption{}
        \label{fig:sub2}
    \end{subfigure}
    
    \caption{Illustrative example of the consistency graph and its affinity matrix for point cloud registration, from \cite{lusk2021clipper}. (a) Blue and red point clouds with putative associations $u_1, \ldots, u_5$. (b) Consistency graph $G$, where edges indicate geometric consistency between associations.}
    \label{fig:clipper}
\end{figure}

\subsection{Graph-Theoretic Data Association}

The goal of the data association pipeline is to perform localization between the current map $M_{\text{veh}}$, computed in real time onboard the agent, and a reference map $M_{\text{ref}}$, collected by an agent during a prior session.

We utilize a geometric data association framework \cite{lusk2021clipper} to compute the largest geometrically consistent set of 3D segment correspondences between $M_{\text{veh}}$ and $M_{\text{ref}}$. Geometric consistency is defined via pairwise distance preservation between associated segments. For instance, if points $\mathbf{a}, \mathbf{b} \in M_{\text{veh}}$ are matched to $\mathbf{a}', \mathbf{b}' \in M_{\text{ref}}$, then the pair is considered consistent if:
\[
\|\mathbf{a} - \mathbf{b}\| \approx \|\mathbf{a}' - \mathbf{b}'\|.
\]
An example of this process is shown in Fig.~\ref{fig:clipper}.

We represent candidate associations as nodes in a graph, where the affinity matrix $A \in \{0,1\}^{n \times n}$ defines pairwise consistency:
\[
A_{u_i, u_j} =
\begin{cases}
1 & \text{if associations } u_i \text{ and } u_j \text{ are consistent,} \\
0 & \text{otherwise.}
\end{cases}
\]
To find the largest mutually consistent set of correspondences, we solve for the densest clique via the following optimization:
\begin{equation}
\begin{aligned}
\max_{u \in \{0,1\}^n} \quad & \frac{u^\top A u}{u^\top u} \\
\text{subject to} \quad & u_i u_j = 0 \quad \text{if } A_{i,j} = 0.
\end{aligned}
\end{equation}

Given the resulting set of geometrically consistent segment correspondences $S = \{ (\mathbf{a}_i, \mathbf{b}_i) \}_{i=1}^k$, where $\mathbf{a}_i \in M_{\text{veh}}$ and $\mathbf{b}_i \in M_{\text{ref}}$, we estimate the rigid transformation that aligns the current vehicle map to the reference map using Arun’s method \cite{arun1987leastsquares}. This produces the optimal rotation $\mathbf{R} \in \text{SO}(3)$ and translation $\mathbf{t} \in \mathbb{R}^3$ minimizing:
\[
\sum_{i=1}^k \left\| \mathbf{b}_i - \left( \mathbf{R} \mathbf{a}_i + \mathbf{t} \right) \right\|^2.
\]
The resulting transformation from the vehicle frame to the reference map frame is:
\[
\mathbf{T}_{\text{veh}}^{\text{ref}} =
\begin{bmatrix}
\mathbf{R} & \mathbf{t} \\
\mathbf{0}^\top & 1
\end{bmatrix} \in \text{SE}(3).
\]

\section{Evaluation}
\subsection{Simulator}

\begin{figure}[b!]
    \centering
    \includegraphics[width=0.95\linewidth]{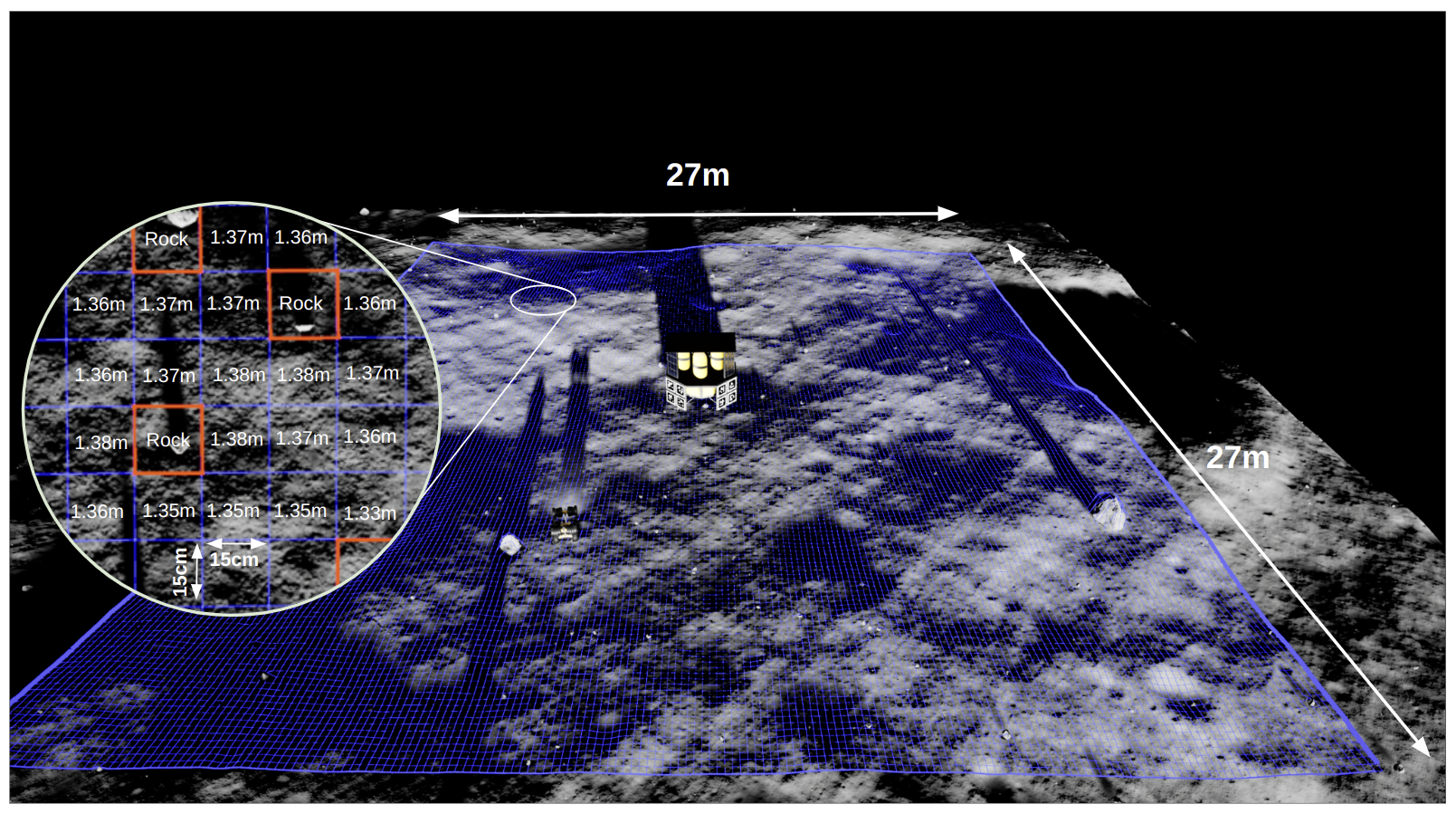}
    \caption{The Lunar Simulator includes a lunar lander at the center of a $27$m x $27$m map, from \cite{lunar_autonomy_challenge}}
    \label{fig:lunarsim}
\end{figure}

We evaluate the performance of our approach using the Lunar Simulator, an extension of the CARLA Simulator \cite{carla2017} developed for the Lunar Autonomy Challenge \cite{lunar_autonomy_challenge}. While CARLA is an open-source platform, the Lunar Simulator adapts its capabilities to replicate the unique challenges of lunar exploration with a high-fidelity, physics-based simulation of the lunar surface, including terrain, lighting conditions, and other environmental factors reflective of space exploration contexts. The Lunar Simulator includes a $27$m x $27$m map with 386 rocks, a lunar lander in the center, and elevation varying across approximately $1$m, as shown in Fig \ref{fig:lunarsim}.

In the Lunar Simulator, we deploy a digital twin of NASA's In-Situ Resource Utilization (ISRU) Pilot Excavator (IPEx) rover. This agent reflects the physical IPEx robot at Kennedy Space Center, which is engineered to excavate and transport lunar regolith. The simulator incorporates detailed models of the IPEx’s sensors, actuators, and overall dynamics. The sensor locations reflect the IPEx rover's design, as shown in Fig. \ref{fig:ipex-diagram}. Cameras are arranged in two sets of stereo-pairs on the front and rear of the rover and as standalone cameras on the sides and excavating drums. All cameras are grayscale and simulated as perfect pinhole cameras.

\subsection{Dataset}

\begin{table}[t]
\centering
\caption{Traverse Data}
\begin{tabular}{@{}cccc@{}}
\toprule

\textbf{Traverse}& \textbf{{Length [m]}} & \textbf{Duration [min]} & \textbf{Object Detections} \\ \midrule

1 & 6.3 & 6 & 1283 \\
2 & 6.5 & 6 & 1280 \\
3 & 5.6 & 6 & 1261 \\
4 & 17.8 & 16 & 3376 \\
5 & 9.6 & 9 & 1884 \\
6 & 8.9 & 6 & 1060 \\
7 & 37.3 & 7 & 6263 \\
8 & 15.6 & 12 & 2374 \\
9 & 54.5 & 10 & 1695 \\
10 & 88.2 & 15 & 2982 \\
11 & 36.5 & 7 & 874 \\
12 & 35.9 & 7 & 871 \\
13 & 197.3 & 26 & 14494 \\
14 & 33.3 & 6 & 2911 \\
15 & 98.3 & 16 & 6722 \\
16 & 8.7 & 2 & 755 \\
17 & 47.2 & 8 & 3701 \\

% 1 & 91.2 & 18 & 4059 \\
% 2 & 41.1 & 11 & 1244 \\
% 3 & 35.9 & 6 & 871 \\

\bottomrule
% \caption{table:rmse}
\label{tab:traverse}
\end{tabular}
\end{table}

\begin{figure}[t]
    \centering
    \includegraphics[width=0.95\linewidth, trim={0, 0, 0, 1.36cm}, clip]{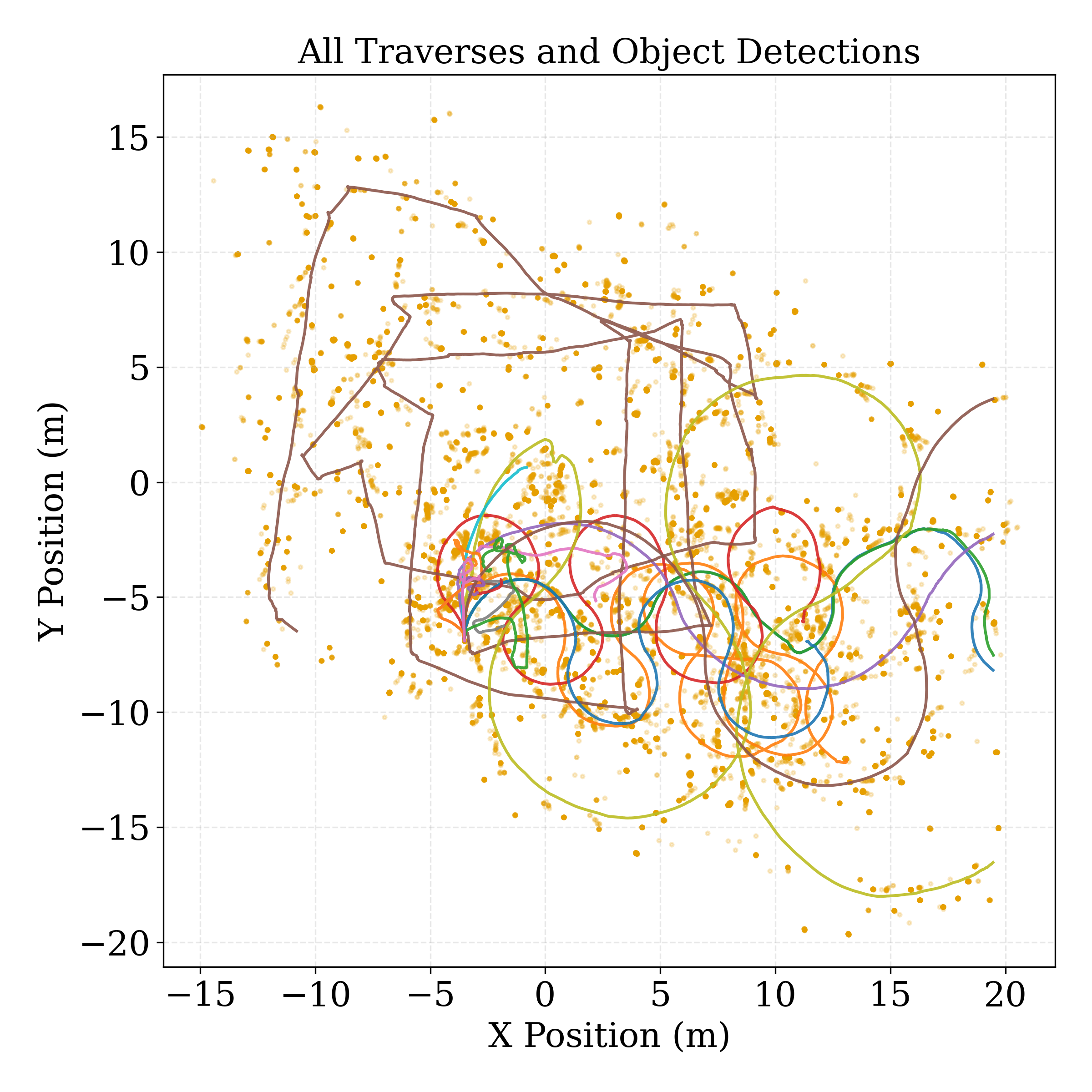}
    \caption{All traverses available in the LunarLoc dataset shown in different colors where detected objects are shown as yellow dots.}
    \label{fig:objects}
\end{figure}

To evaluate the performance of LunarLoc, we collected stereo images and poses from short, medium, and long traverses spanning the same general region of the simulator map covering an area of approximately $1000$ square meters.
All 17 traverses are shown in Fig. \ref{fig:objects}, and detailed information on each traverse is provided in Table \ref{tab:traverse}. 
Subsets of longer traverses are taken to form shorter traverses for analysis.
Each traverse is made available as a simple comma-separated table consisting of the IPEx position and boulder detections in each frame and as a \texttt{.lac} archive file, containing more detailed sensor information and time synchronized image data, details of which are outlined in Table \ref{tab:metrics}.

\begin{table}[b]
\centering
\caption{Metrics recorded in each LAC archive}
\begin{tabular}{@{}ll@{}}
\toprule
\textbf{Sensor Group} & \textbf{Recorded Metrics (frequency)} \\
\midrule
\textbf{Rover State @ 20Hz} &
\begin{tabular}[t]{@{}l@{}}
Mission Time \\
Accelerometer Measurements \\
Gyroscope Measurements \\
Power Consumption \\
Control Input (linear and angular velocity) \\
Rover Pose (position and orientation)
\end{tabular} \\
\addlinespace
\textbf{Camera @ up to 10Hz} &
\begin{tabular}[t]{@{}l@{}}
Camera Images (subsampled to 0.5Hz) \\
Camera Enable / Disable State \\
Camera Light Intensity \\
Camera / Light Position
\end{tabular} \\
\bottomrule
\end{tabular}
\label{tab:metrics}
\end{table}

\begin{figure*}[t!]
    \centering
    % First subfigure
    \begin{subfigure}{0.3\textwidth}
        \includegraphics[width=\linewidth,  trim={0, 0, 0, 1cm}, clip]{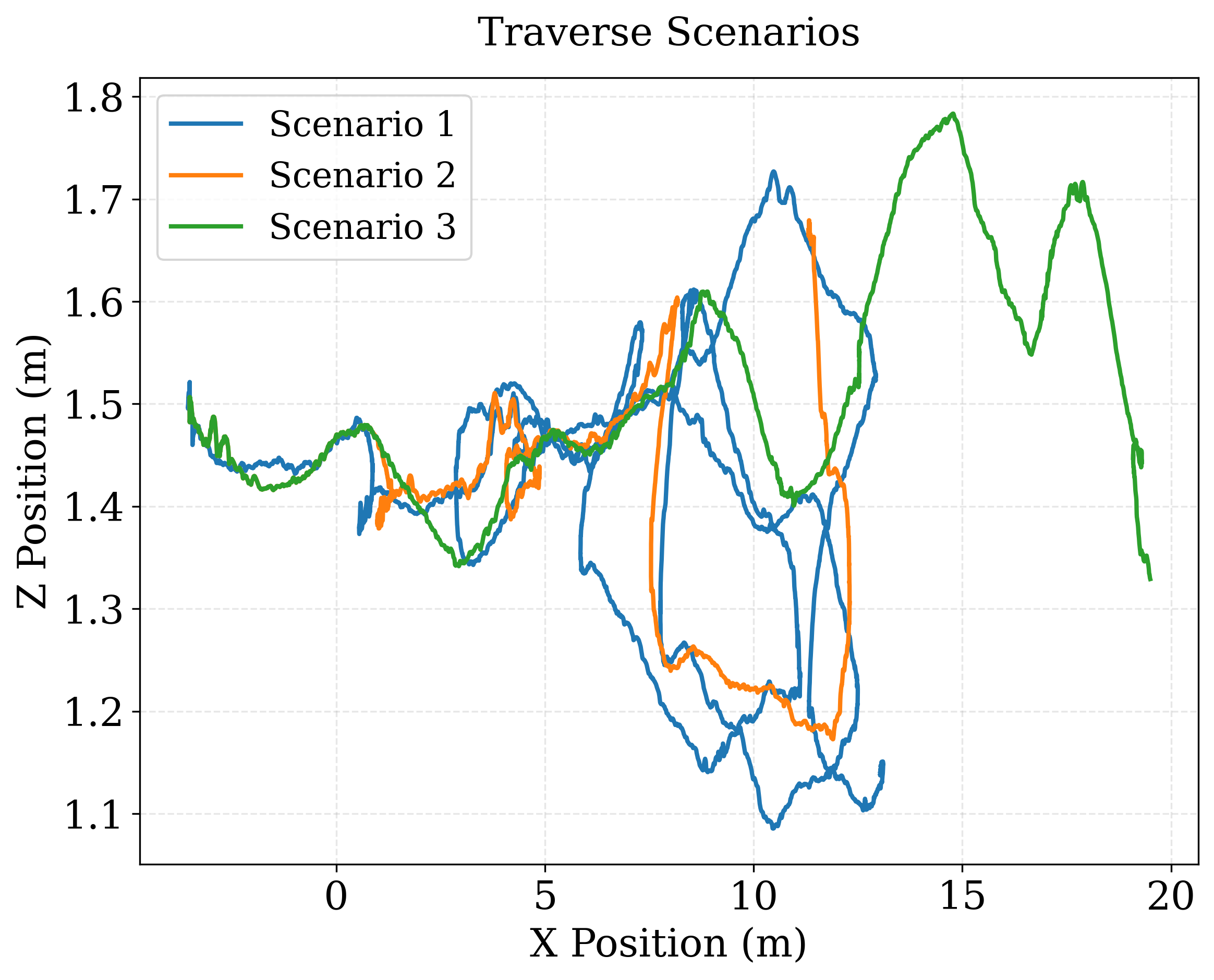}
        \caption{X-Z Traverse}
        \label{fig:sub1}
    \end{subfigure}
    \hfill
    % Second subfigure
    \begin{subfigure}{0.3\textwidth}
        \includegraphics[width=\linewidth,  trim={0, 0, 0, 1cm}, clip]{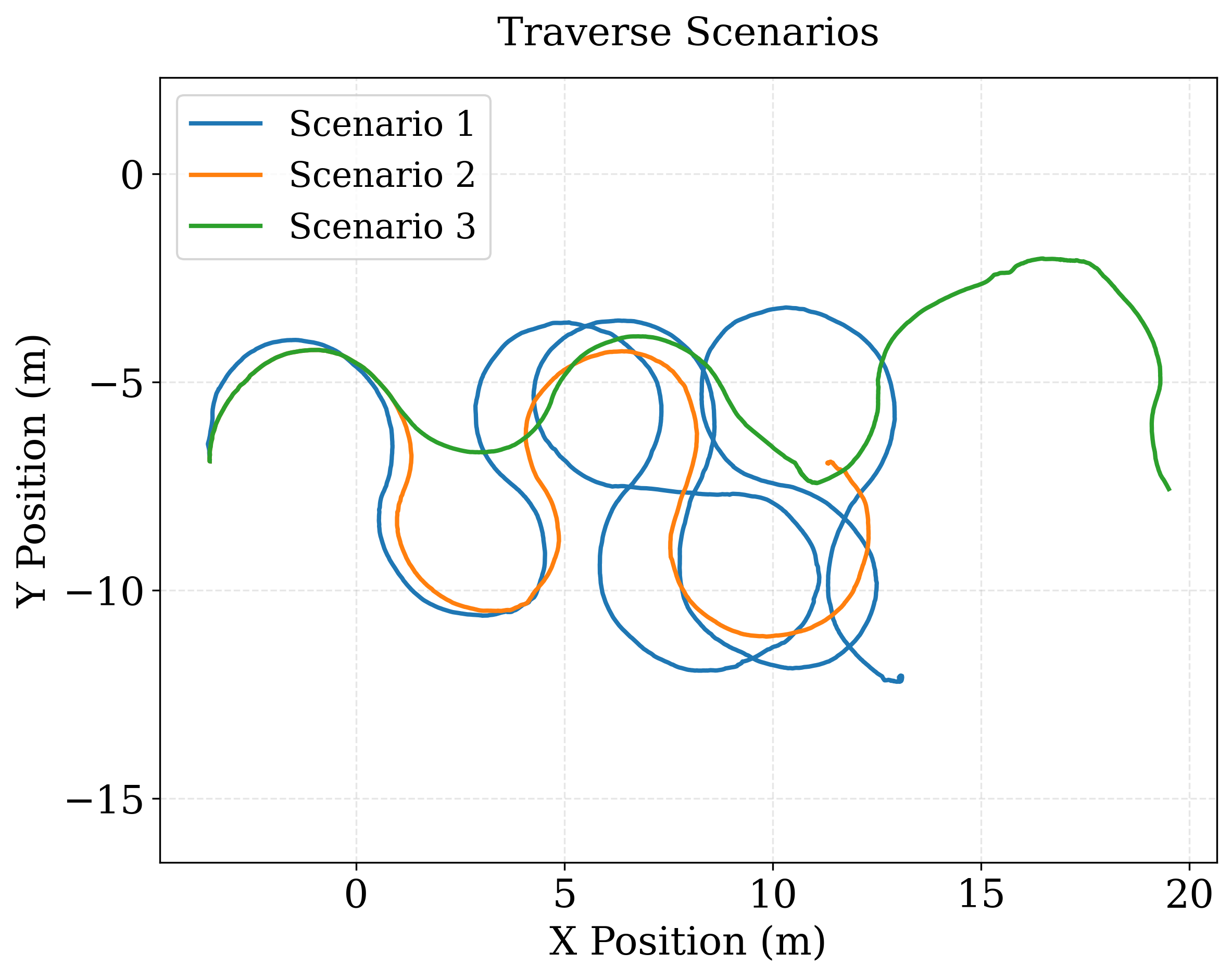}
        \caption{X-Y Traverse}
        \label{fig:sub2}
    \end{subfigure}
    \hfill
    % Third subfigure
    \begin{subfigure}{0.3\textwidth}
        \includegraphics[width=\linewidth,  trim={0, 0, 0, 1cm}, clip]{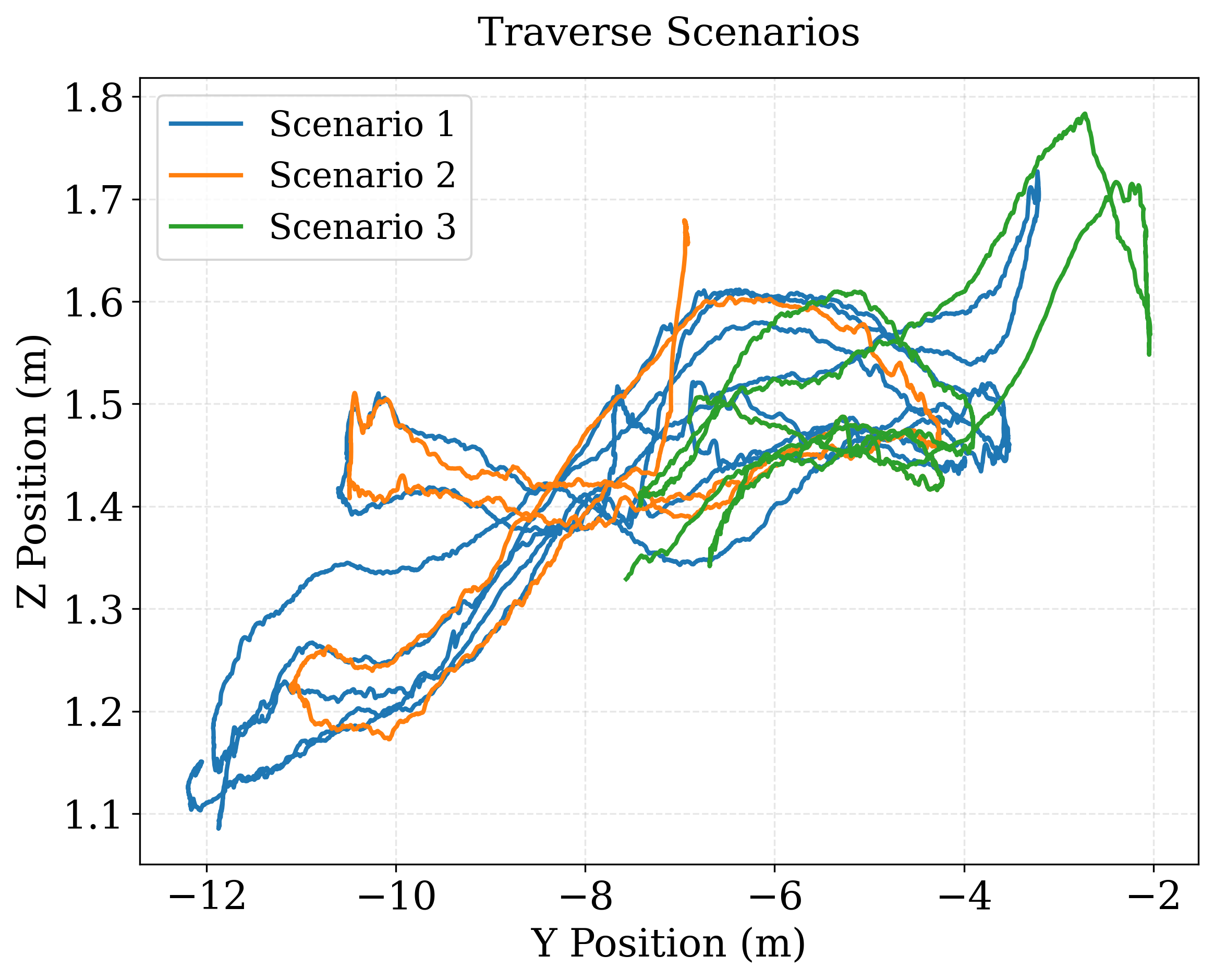}
        \caption{Y-Z Traverse}
        \label{fig:sub3}
    \end{subfigure}
    
    \caption{Ground truth traverses from the Lunar Simulator used to evaluate LunarLoc. Scenarios 1, 2 and 3 refer to traverses used for multi-session global localization.}
    \label{fig:traverses}
\end{figure*}

Each \texttt{.lac} archive is a gzip tarball.
For ease of use, a Python library is provided to access synchronized camera and sensor data.
The library provides direct access to underlying tabular data as well as an API mimicking the sensor interfaces provided by the Lunar Simulator.
This mock API enables the playback of previous traverses outside of the simulator and without any reliance on the simulator code.
Localization methods developed using the data loader can be quickly tested outside of the resource intensive simulator and then seamlessly deployed in the Lunar Simulator, accelerating development.
Traverse data and the python data loader utility are made available on GitHub at \url{https://github.com/mit-acl/lunarloc-data}.

% \begin{figure}[b]
%     \centering
% \begin{verbatim}
% <file_name>.lac/
% ├─ metadata.toml 
% ├─ initial.toml 
% ├─ frames.csv 
% ├─ images/ 
% │  └─ <camera>/ 
% │     ├─ <camera>_frames.csv 
% │     ├─ grayscale/ 
% │     │  └─ <camera>_grayscale_<frame>.png 
% │     └─ semantic/ 
% │        └─ <camera>_semantic_<frame>.png 
% └─ custom/ 
%    └─ <record_name>.csv
% \end{verbatim}
%     \caption{\texttt{.lac} Archive Format}
%     \label{fig:lac-format}
% \end{figure}

To collect data for each traverse, the IPEx is initialized at a common starting location from which it navigates to randomly selected way points in the immediate region until the minimum path length for the traverse is reached.
This exploration strategy increases the frequency of repeated observations of rocks and other terrain features within and between traverses.
Images from the front-left and front-right stereo cameras are sampled at $5$Hz and used as inputs for the zero-shot segmentation and object centroid estimation pipeline.
If suitable objects are detected in the scene the estimated position of the object's centroid is recorded along with the rover's ground truth position and other state information.
This information is saved as an individual file for each traverse enabling post processing and performance comparison of localization methods outside of the resource intensive simulator.
Through the use of the provided playback utility, it is possible to synchronize multiple separate traverses, extending beyond multi-session localization to model multi-agent systems operating in the same environment.

% Discuss the collected traverses here - basically that they are small, medium, and long traverses and that we include a fourth which is just a subset of the longest one (so we don't evaluate that one against itself).

\section{Experimental Results}

We evaluate the performance of LunarLoc using five pairs of traverses from the Lunar Simulator shown in Fig. \ref{fig:traverses}, including all combinations of four traverses aside from the longest traverse with the shorter portion of itself, as Traverse 3 is a subset of Traverse 12. We report the number of segments found in each submap as well as the number of inlier associations returned during data association in Table \ref{table:rmse}. We deploy LunarLoc in real time. On a NVIDIA GeForce RTX 3090, the mean detection time of FastSAM runs in real time at 25 Hz \cite{zhao2023fast} which is the slowest component of LunarLoc's front-end detection module. 

LunarLoc uses the underlying structure of the environment, i.e. boulder locations, as a cue for global localization via a graph theoretic approach. With the number of segments detected ranging from 36 to 185 with up to 41\% inlier associations, we show that rock detections are a reliable cue for data association in this setting.

\begin{table}[h]
\centering
\caption{Tracking Performance on LunarLoc Dataset (RMSE ↓ [m]). }
\begin{tabular}{@{}cccccccccc@{}}
\toprule

\textbf{Path 1}& \textbf{{Path 2}} & \textbf{Segs in 1} & \textbf{Segs in 2} & \textbf{Ain} & \textbf{RMSE ↓ [cm]} \\ \midrule

3 & 5 & 36 & 62 & 15 & 0.08  \\
3 & 7 & 36 & 102 & 22 & 0.61  \\
5 & 7 & 62 & 102 & 28 & 1.79  \\
5 & 12 & 62 & 185 & 18 & 0.32  \\
7 & 12 & 102 & 185 & 41 & 0.68  \\

\bottomrule
\label{table:rmse}
\end{tabular}
\end{table}

To evaluate the localization performance of LunarLoc, we report the tracking performance defined by the root mean squared error of the translation component $\mathbf{t}$ between the vehicle map $M_{\text{veh}}$ and the reference map $M_{\text{ref}}$. Transformations returned by LunarLoc as shown in Table \ref{table:rmse} have localization error less than $2$ cm in all evaluated cases, significantly outperforming the state of the art for localization on the Moon as shown in Table \ref{tab:planetary-localization}.

\section{Conclusion} 
\label{sec:conclusion}

We propose LunarLoc, a global localization method using graph-theoretic data association of objects in the scene extracted using zero-shot segmentation. Future work includes integrating this localization framework into multi-agent settings and performing more extensive testing from different viewpoints and varying lighting conditions. We also plan to perform global localization with respect to an aerial map of the rock locations taken during landing. While place recognition approaches struggle with perceptual aliasing and feature-based methods may not be distinct enough to operate in various lighting conditions \cite{10801471}, LunarLoc is a promising approach using the underlying structure of the terrain as a cue for global localization on the Moon.

% \section*{Acknowledgments}

% The simulation used for this research was developed as part of the Lunar Autonomy Challenge. The Lunar Autonomy Challenge is a collaboration between NASA, The Johns Hopkins University (JHU) Applied Physics Laboratory (APL), Caterpillar Inc., and Embodied AI. APL is managing the challenge for NASA.

%% Use plainnat to work nicely with natbib. 

% \bibliographystyle{plainnat}
% \bibliography{references}

\begin{thebibliography}{30}
\providecommand{\natexlab}[1]{#1}
\providecommand{\url}[1]{#1}
\csname url@samestyle\endcsname
\providecommand{\newblock}{\relax}
\providecommand{\bibinfo}[2]{#2}
\providecommand{\BIBentrySTDinterwordspacing}{\spaceskip=0pt\relax}
\providecommand{\BIBentryALTinterwordstretchfactor}{4}
\providecommand{\BIBentryALTinterwordspacing}{\spaceskip=\fontdimen2\font plus
\BIBentryALTinterwordstretchfactor\fontdimen3\font minus \fontdimen4\font\relax}
\providecommand{\BIBforeignlanguage}[2]{{%
\expandafter\ifx\csname l@#1\endcsname\relax
\typeout{** WARNING: IEEEtranN.bst: No hyphenation pattern has been}%
\typeout{** loaded for the language `#1'. Using the pattern for}%
\typeout{** the default language instead.}%
\else
\language=\csname l@#1\endcsname
\fi
#2}}
\providecommand{\BIBdecl}{\relax}
\BIBdecl

\bibitem[Arandjelovic et~al.(2016)Arandjelovic, Gronat, Torii, Pajdla, and Sivic]{arandjelovic2016netvlad}
R.~Arandjelovic, P.~Gronat, A.~Torii, T.~Pajdla, and J.~Sivic, ``Netvlad: Cnn architecture for weakly supervised place recognition,'' in \emph{Proceedings of the IEEE conference on computer vision and pattern recognition}, 2016, pp. 5297--5307.

\bibitem[Oquab et~al.(2023)Oquab, Darcet, Moutakanni, Vo, Szafraniec, Khalidov, Fernandez, Haziza, Massa, El-Nouby, Howes, Huang, Xu, Sharma, Li, Galuba, Rabbat, Assran, Ballas, Synnaeve, Misra, Jegou, Mairal, Labatut, Joulin, and Bojanowski]{oquab2023dinov2}
M.~Oquab, T.~Darcet, T.~Moutakanni, H.~V. Vo, M.~Szafraniec, V.~Khalidov, P.~Fernandez, D.~Haziza, F.~Massa, A.~El-Nouby, R.~Howes, P.-Y. Huang, H.~Xu, V.~Sharma, S.-W. Li, W.~Galuba, M.~Rabbat, M.~Assran, N.~Ballas, G.~Synnaeve, I.~Misra, H.~Jegou, J.~Mairal, P.~Labatut, A.~Joulin, and P.~Bojanowski, ``Dinov2: Learning robust visual features without supervision,'' 2023.

\bibitem[Lowe(2004)]{sift2004}
D.~Lowe, ``Distinctive image features from scale-invariant keypoints,'' \emph{International Journal of Computer Vision}, vol.~60, pp. 91--, 11 2004.

\bibitem[Rublee et~al.(2011)Rublee, Rabaud, Konolige, and Bradski]{orb2011}
E.~Rublee, V.~Rabaud, K.~Konolige, and G.~Bradski, ``Orb: An efficient alternative to sift or surf,'' in \emph{2011 International Conference on Computer Vision}, 2011, pp. 2564--2571.

\bibitem[Fischler and Bolles(1981)]{fischler1981random}
M.~A. Fischler and R.~C. Bolles, ``Random sample consensus: a paradigm for model fitting with applications to image analysis and automated cartography,'' \emph{Communications of the ACM}, vol.~24, no.~6, pp. 381--395, 1981.

\bibitem[Thomas et~al.(2024)Thomas, Kinnari, Lusk, Kondo, and How]{10801471}
A.~Thomas, J.~Kinnari, P.~C. Lusk, K.~Kondo, and J.~P. How, ``Sos-match: Segmentation for open-set robust correspondence search and robot localization in unstructured environments,'' in \emph{2024 IEEE/RSJ International Conference on Intelligent Robots and Systems (IROS)}, 2024, pp. 5613--5620.

\bibitem[Zhao et~al.(2023)Zhao, Ding, An, Du, Yu, Li, Tang, and Wang]{zhao2023fast}
X.~Zhao, W.~Ding, Y.~An, Y.~Du, T.~Yu, M.~Li, M.~Tang, and J.~Wang, ``Fast segment anything,'' 2023.

\bibitem[Cummins and Newman(2008)]{cummins2008fab}
M.~Cummins and P.~Newman, ``Fab-map: Probabilistic localization and mapping in the space of appearance,'' \emph{The International journal of robotics research}, vol.~27, no.~6, pp. 647--665, 2008.

\bibitem[Sarlin et~al.(2019)Sarlin, Cadena, Siegwart, and Dymczyk]{sarlin2019coarse}
P.-E. Sarlin, C.~Cadena, R.~Siegwart, and M.~Dymczyk, ``From coarse to fine: Robust hierarchical localization at large scale,'' in \emph{Proceedings of the IEEE/CVF conference on computer vision and pattern recognition}, 2019, pp. 12\,716--12\,725.

\bibitem[Keetha et~al.(2023)Keetha, Mishra, Karhade, Jatavallabhula, Scherer, Krishna, and Garg]{keetha2023anyloc}
N.~Keetha, A.~Mishra, J.~Karhade, K.~M. Jatavallabhula, S.~Scherer, M.~Krishna, and S.~Garg, ``Any{L}oc: Towards universal visual place recognition,'' \emph{arXiv preprint arXiv:2308.00688}, 2023.

\bibitem[DeTone et~al.(2018)DeTone, Malisiewicz, and Rabinovich]{detone2018superpoint}
D.~DeTone, T.~Malisiewicz, and A.~Rabinovich, ``Superpoint: Self-supervised interest point detection and description,'' in \emph{Proceedings of the IEEE conference on computer vision and pattern recognition workshops}, 2018, pp. 224--236.

\bibitem[Grimes and LeCun(2009)]{grimes2009efficient}
M.~Grimes and Y.~LeCun, ``Efficient off-road localization using visually corrected odometry,'' in \emph{2009 IEEE International Conference on Robotics and Automation}.\hskip 1em plus 0.5em minus 0.4em\relax IEEE, 2009, pp. 2649--2654.

\bibitem[Ort et~al.(2018)Ort, Paull, and Rus]{ort2018autonomous}
T.~Ort, L.~Paull, and D.~Rus, ``Autonomous vehicle navigation in rural environments without detailed prior maps,'' in \emph{2018 IEEE international conference on robotics and automation (ICRA)}.\hskip 1em plus 0.5em minus 0.4em\relax IEEE, 2018, pp. 2040--2047.

\bibitem[Ankenbauer et~al.(2023)Ankenbauer, Lusk, and How]{ankenbauer2023global}
J.~Ankenbauer, P.~C. Lusk, and J.~P. How, ``Global localization in unstructured environments using semantic object maps built from various viewpoints,'' \emph{arXiv preprint arXiv:2303.04658}, 2023.

\bibitem[Redmon and Farhadi(2018)]{redmon2018yolov3}
J.~Redmon and A.~Farhadi, ``Yolov3: An incremental improvement,'' \emph{arXiv preprint arXiv:1804.02767}, 2018.

\bibitem[Peterson et~al.(2024)Peterson, Jia, Tian, Thomas, and How]{peterson2024roman}
M.~B. Peterson, Y.~X. Jia, Y.~Tian, A.~Thomas, and J.~P. How, ``Roman: Open-set object map alignment for robust view-invariant global localization,'' \emph{arXiv preprint arXiv:2410.08262}, 2024.

\bibitem[Singh and Leonard(2024)]{singh2024open}
K.~Singh and J.~J. Leonard, ``Open-set semantic uncertainty aware metric-semantic graph matching,'' \emph{arXiv preprint arXiv:2409.11555}, 2024.

\bibitem[{Parker} et~al.(2010){Parker}, {Golombek}, and {Powell}]{parker2010geomorphic}
T.~J. {Parker}, M.~P. {Golombek}, and M.~W. {Powell}, ``{Geomorphic/Geologic Mapping, Localization, and Traverse Planning at the Opportunity Landing Site, Mars},'' in \emph{41st Annual Lunar and Planetary Science Conference}, ser. Lunar and Planetary Science Conference, Mar. 2010, p. 2638.

\bibitem[Maimone et~al.(2006)Maimone, Biesiadecki, Tunstel, Cheng, and Leger]{maimone2006surface}
M.~Maimone, J.~Biesiadecki, E.~Tunstel, Y.~Cheng, and C.~Leger, ``Surface navigation and mobility intelligence on the mars exploration rovers. in intelligence for space robotics,'' 01 2006.

\bibitem[Johnson et~al.()Johnson, Ansar, Matthies, Trawny, Mourikis, and Roumeliotis]{johnson2007general}
\BIBentryALTinterwordspacing
A.~Johnson, A.~Ansar, L.~Matthies, N.~Trawny, A.~Mourikis, and S.~Roumeliotis, \emph{A General Approach to Terrain Relative Navigation for Planetary Landing}. [Online]. Available: \url{https://arc.aiaa.org/doi/abs/10.2514/6.2007-2854}
\BIBentrySTDinterwordspacing

\bibitem[Hook et~al.(2022)Hook, Schwartz, Ebadi, Coble, and Padgett]{hook2022topographical}
J.~V. Hook, R.~Schwartz, K.~Ebadi, K.~Coble, and C.~Padgett, ``Topographical landmarks for ground-level terrain relative navigation on mars,'' in \emph{2022 IEEE Aerospace Conference (AERO)}, 2022, pp. 1--6.

\bibitem[van Diggelen and Enge(2015)]{vanDiggelen2015gpsmooc}
F.~van Diggelen and P.~Enge, ``The world’s first gps mooc and worldwide laboratory using smartphones,'' in \emph{Proceedings of the 28th International Technical Meeting of the Satellite Division of The Institute of Navigation (ION GNSS+ 2015)}, Tampa, Florida, Sep. 2015, pp. 361--369.

\bibitem[San~Martin et~al.(2017)San~Martin, Bayard, Conway, Mandic, and Bailey]{martin2017maven}
\BIBentryALTinterwordspacing
M.~A. San~Martin, D.~S. Bayard, D.~T. Conway, M.~Mandic, and E.~S. Bailey, ``{A Minimal State Augmentation Algorithm for Vision-Based Navigation without Using Mapped Landmarks},'' 2017. [Online]. Available: \url{https://hdl.handle.net/2014/46668}
\BIBentrySTDinterwordspacing

\bibitem[Grip et~al.(2022)Grip, Conway, Lam, Williams, Golombek, Brockers, Mischna, and Cacan]{grip2022ingenuity}
H.~F. Grip, D.~Conway, J.~Lam, N.~Williams, M.~P. Golombek, R.~Brockers, M.~Mischna, and M.~R. Cacan, ``Flying a helicopter on mars: How ingenuity's flights were planned, executed, and analyzed,'' in \emph{2022 IEEE Aerospace Conference (AERO)}, 2022, pp. 1--17.

\bibitem[Daftry et~al.(2023)Daftry, Chen, Cheng, Tepsuporn, Coltin, Naam, Ma, Khattak, Deans, and Matthies]{daftry2023lunarnav}
\BIBentryALTinterwordspacing
S.~Daftry, Z.~Chen, Y.~Cheng, S.~Tepsuporn, B.~Coltin, U.~Naam, L.~M. Ma, S.~Khattak, M.~Deans, and L.~Matthies, ``Lunarnav: Crater-based localization for long-range autonomous lunar rover navigation,'' 2023. [Online]. Available: \url{https://arxiv.org/abs/2301.01350}
\BIBentrySTDinterwordspacing

\bibitem[Atha et~al.(2024)Atha, Michael~Swan, Cauligi, Bettens, Goh, Kogan, Matthies, and Ono]{atha2024shadownav}
\BIBentryALTinterwordspacing
D.~Atha, R.~Michael~Swan, A.~Cauligi, A.~Bettens, E.~Goh, D.~Kogan, L.~Matthies, and M.~Ono, ``Shadownav: Autonomous global localization for lunar navigation in darkness,'' \emph{IEEE Transactions on Field Robotics}, vol.~1, p. 213–230, 2024. [Online]. Available: \url{http://dx.doi.org/10.1109/TFR.2024.3462391}
\BIBentrySTDinterwordspacing

\bibitem[{Johns Hopkins University Applied Physics Laboratory}(2024)]{lunar_autonomy_challenge}
\BIBentryALTinterwordspacing
{Johns Hopkins University Applied Physics Laboratory}, ``Lunar autonomy challenge documentation,'' 2024, accessed: 2025-05-15. [Online]. Available: \url{https://lunar-autonomy-challenge.jhuapl.edu/Challenge-Documentation/index.php#introduction}
\BIBentrySTDinterwordspacing

\bibitem[Lusk et~al.(2021)Lusk, Fathian, and How]{lusk2021clipper}
P.~C. Lusk, K.~Fathian, and J.~P. How, ``Clipper: A graph-theoretic framework for robust data association,'' in \emph{2021 IEEE International Conference on Robotics and Automation (ICRA)}.\hskip 1em plus 0.5em minus 0.4em\relax IEEE, 2021, pp. 13\,828--13\,834.

\bibitem[Arun et~al.(1987)Arun, Huang, and Blostein]{arun1987leastsquares}
K.~S. Arun, T.~S. Huang, and S.~D. Blostein, ``Least-squares fitting of two 3-{D} point sets,'' \emph{IEEE Transactions on Pattern Analysis and Machine Intelligence}, vol. PAMI-9, no.~5, pp. 698--700, 1987.

\bibitem[Dosovitskiy et~al.(2017)Dosovitskiy, Ros, Codevilla, Lopez, and Koltun]{carla2017}
A.~Dosovitskiy, G.~Ros, F.~Codevilla, A.~Lopez, and V.~Koltun, ``{CARLA}: {An} open urban driving simulator,'' in \emph{Proceedings of the 1st Annual Conference on Robot Learning}, 2017, pp. 1--16.

\end{thebibliography}
\bibliographystyle{plainnat}  % or whatever you're using
% Generated by IEEEtranN.bst, version: 1.14 (2015/08/26)

              % matches your .bbl file name

\end{document}